\documentclass[10pt,twocolumn,letterpaper]{article}

\usepackage{ijcb}
\usepackage{times}
\usepackage{epsfig}
\usepackage{graphicx}
\usepackage{amsmath}
\usepackage{amssymb}
\usepackage{multirow}
\usepackage{multicol}

\usepackage{color, colortbl}
\definecolor{LightCyan}{rgb}{0.88,1,1}

\ijcbfinalcopy 

\usepackage{algorithm}
\usepackage{algpseudocode}
\usepackage{mathtools}
\usepackage{pifont}
\usepackage{booktabs}
\usepackage{todonotes}
\usepackage[bb=boondox]{mathalfa}
\usepackage{multirow}

\DeclareMathOperator*{\argmin}{arg\,min} 
\algnewcommand{\LeftComment}[1]{\Statex \(\triangleright\) #1}

\ifijcbfinal\fi

\begin{document}

\title{GaitMorph: Transforming Gait by Optimally Transporting Discrete Codes}

\author{Adrian Cosma, Emilian Rădoi\\
University Politehnica of Bucharest, Bucharest, Romania\\
{\tt\small cosma.i.adrian@gmail.com,  emilian.radoi@upb.ro}
}

\maketitle
\thispagestyle{empty}

\begin{abstract} 
   Gait, the manner of walking, has been proven to be a reliable biometric with uses in surveillance, marketing and security. A promising new direction for the field is training gait recognition systems without explicit human annotations, through self-supervised learning approaches. Such methods are heavily reliant on strong augmentations for the same walking sequence to induce more data variability and to simulate additional walking variations. Current data augmentation schemes are heuristic and cannot provide the necessary data variation as they are only able to provide simple temporal and spatial distortions. In this work, we propose GaitMorph, a novel method to modify the walking variation for an input gait sequence. Our method entails the training of a high-compression model for gait skeleton sequences that leverages unlabelled data to construct a discrete and interpretable latent space, which preserves identity-related features. Furthermore, we propose a method based on optimal transport theory to learn latent transport maps on the discrete codebook that morph gait sequences between variations. We perform extensive experiments and show that our method is suitable to synthesize additional views for an input sequence.
\end{abstract}

\section{Introduction}
The way people walk, also known as gait, is a crucial biometric trait that has numerous applications in medicine \cite{medical-gait}, sports \cite{sports-gait}, and surveillance\cite{10.1145/3490235}. Most notably, in recent years, it has been successfully used as a unique biometric fingerprint to accurately identify individuals from a distance \cite{cosma22gaitformer}. 
The biggest challenge in gait analysis \cite{gait-survey} is disentangling confounding factors which significantly affect and obfuscate gait, such as the individual clothing, footwear, walking speed, injury, state of mind, and social environment. Moreover the extrinsic characteristics of gait sensors (such as camera viewpoint, distance and resolution) severely affect the quality of the captured gait. Developing a robust model, able to ignore these factors and represent the essential gait characteristics is still an open problem. Consequently, deploying highly accurate gait recognition systems in real-world unconstrained scenarios remains a difficult problem.

Previous works \cite{cosma22gaitformer,gait-vit} have shown that self-supervised pretraining is a promising new direction, but is still not enough to achieve high performance modelling. Self-supervised pre-training exposes the backbone model to a large variety of walking registers, increasing the robustness in downstream tasks. However, contrastive pre-training requires high degree of variation in the data \cite{chen2020simple,tian2020makes}, which is often hard to obtain automatically for gait. Fine-tuning is still necessary for effective gait recognition \cite{gait-vit}, especially in specific and uncommon environments. Currently, data variation for gait modelling systems is obtained by using data augmentation techniques \cite{cosma22gaitformer}, which have the goal to distort the gait sequence while preserving the person identity. However, heuristical augmentation procedures are not able to reliably produce novel viewpoints for a gait sequence, or to seamlessly change the walking variation as they only provide simple temporal and spatial distortions. For other similar tasks such as person re-identification \cite{reid-survey}, viewpoint variation is induced through learned methods such as approaches in human pose transfer \cite{sanyal2021learning}. 

We propose GaitMorph, a novel method that is able to modify skeleton gait sequences to synthesize novel views. We use a high-compression model for gait sequences that leverages large amounts of in-the-wild and unlabelled data to construct a discrete and interpretable latent space for skeleton gait sequences. Our model is based on the vector-quantized variational autoencoder (VQ-VAE) \cite{van2017neural}, and achieves a high degree of compression for skeleton sequences (up to 500$\times$ lower storage demands). We show that the model is able to reconstruct gait sequences with a high degree of fidelity, without losing identity-related features.

Furthermore, compressing gait sequences in a discrete latent space enables easy manipulation of codebook entries between walking variations. We propose to make use of optimal transport \cite{villani2009optimal} to learn transport maps between walking variations, allowing morphing gait sequences into a desired variation or viewpoint.

This work makes the following contributions:

\begin{enumerate}
    \item We demonstrate that compression of gait sequences into a discrete latent space is feasible, and can be achieved while preserving identity-related information of the underlying walker. We achieve a high degree (500$\times$) of compression without deteriorating downstream gait recognition performance (maximum of 3\% accuracy loss for normal walking). 
    
    \item We propose a novel method to morph gait sequences between variations. Using optimal transport theory, we learn transport maps between variations that generate realistic and novel views for a walk. Our experiments show that the distribution of morphed sequences is similar to the real walk distribution, potentially making our method useful for data augmentation.
    
    \item We perform extensive experiments on the core aspect of our proposed method: the VQ-VAE dictionary size. We show that, while a small dictionary size obtains good reconstruction error at a high compression level, the latent space is not sufficiently disentangled to allow easy morphing.
\end{enumerate}

\section{Related Work}
Works in motion sythetisation are predominantly directed towards generating controllable, general actions for use in animation \cite{siyao2022bailando,li2022skeleton2humanoid,raab2023single,cai2021unified}. Yan et al. \cite{yan2019convolutional} proposed a convolutional architecture named Convolutional Sequence Generation Network (CSGN) for generating skeleton sequences for action recognition. The authors employed spatial graph downsampling and temporal downsampling to generate the whole sequence in a single pass, using latent vectors sampled from gaussian processes. Petrovich \cite{petrovich2021action} employed a transformer VAE model conditioned on the action. 

Li et. al \cite{li2022skeleton2humanoid} proposed a method for performing motion "in-betweening" using physically plausible constraints. Raab et al. \cite{raab2023single} perform motion in-betweening by using diffusion models. Wang et al. \cite{wang2022towards} constructed a method for generating movement animations which also takes the target environment into account.

Some works tackle the problem of motion prediction \cite{guo2023back,ma2022progressively,mao2019learning}. Ma et al. \cite{ma2022progressively} used a graph-convolutional network for motion prediction of skeleton sequences. Zhang et al. \cite{zhang2020perpetual} generate unbounded motion sequences conditioned only on a single starting skeleton. The authors employ an RNN-based architecture to procedurally generate skeletons. 

Motion generation techniques have also been used for sign language generation \cite{liu2022learning,xie2022vector}. Liu et al. \cite{liu2022learning} used a cross-modal approach for audio to sign pose sequence generation using a GRU-based model. Xie et al. \cite{xie2022vector} used a VQ-VAE to generate sign pose sequences, using a discrete diffusion prior model. Zhang et al. \cite{zhang2023t2m} propose a Motion VQ-VAE for text-conditioned action generation, and demonstrate that a simple VQ-VAE recipe \cite{razavi2019generating} can have very good performance for this data modality without any major bells and whistles. 

In the area of gait recognition, synthesising walks has been only briefly studied in the past, partially due to the lack of large-scale datasets, and the unique constraints of this settings. Works in self-supervised for images\cite{chen2020simple,guo2022contrastive,tian2020makes} point out that the high quality data augmentation is crucial for learning good representations. Tian et al.\cite{tian2020makes} argues that optimal views for self-supervised contrastive learning are task-dependent. For instance, in gait analysis, Yu et al. \cite{gaitgan} train a generative adversarial network to generate silhouette sequences that are invariant to walking confounding factors such as viewpoint and clothing change. However, the goal was downstream identification and not generation in itself. Yao et al. \cite{app13042084} propose a framework for walking synthetisation based on an autoencoder and a parametric body model, but their experiments are mainly based on silhouette-based identification models.  Different from previous works, we are interested in manipulating the walking variation and viewpoint of existing walks.

\section{Method}
The use of a VQ-VAE \cite{van2017neural} for learning a latent walking representation for skeletons is motivated by the discrete nature of the latent embeddings, which simplifies the constraint optimization for morphing between walking variations. While the VQ-VAE is widely used in generative modelling \cite{zhang2023t2m,xie2022vector,liu2022learning}, other algorithms such as diffusion models \cite{zhang2022motiondiffuse} might offer higher quality reconstructions. However, our aim is not to generate walking sequences, but to manipulate the latent space to change existing walks into desired variations.

In this section, we describe the main components of GaitMorph: we describe the pretraining dataset used for training the VQ-VAE, the architecture and training procedure for the VQ-VAE, and the morphing algorithm based on optimal transport between the latent codes. 

\subsection{Pretraining Dataset}

\label{sec:dataset}

In order to train a sufficiently large and general autoencoder model, we assess that current gait datasets are too small. Even though datasets such as DenseGait \cite{cosma22gaitformer} and GREW \cite{grew} are collected "in-the-wild" outdoor environments using surveillance cameras, they nonetheless lack some walking registers such as treadmill walking, more aggressive camera angles and indoor environments. However, by combining the major large-scale gait datasets into a single dataset, we can ensure more diversity of walking registers. In Table \ref{tab:walkpile} we showcase the existing datasets that comprises our pretraining dataset. We used \textbf{DenseGait} \cite{cosma22gaitformer} and \textbf{GREW} \cite{grew}, two similar in-the-wild datasets for their diverse walking sequences in outdoor environments, \textbf{OU-ISIR} \cite{ouisir} for more controlled walking in indoor and treadmill registers, and \textbf{Gait3D} \cite{zheng2022gait3d}, and indoor "in-the-wild" dataset collected in a supermarket setting. After concatenation of all skeleton sequences from the datasets, we obtain 875,543 walking sequences, totalling 1220.06 hours. To increase the size as much as possible, we also included the testing / distractor splits of each dataset whenever possible. We purposely did not include controlled, small scale datasets such as CASIA-B \cite{casia}, as we use them for downstream evaluation.

All walking sequences in this dataset are 2D skeletons in COCO pose format. We chose 2D poses to unify all datasets, as every dataset is providing 2D poses by default, while only some are also providing silhouettes or body meshes. Even though many gait processing models have good results using and appearance-based approach with silhouettes \cite{chao2019gaitset,lin2022gaitgl}, pose sequences only encode movement and abstract away any appearance information, preserving the privacy of walking individuals \cite{gaitgraph,cosma22gaitformer}. Skeleton sequences are a more interpretable and a plethora of models employ them for motion synthesis \cite{siyao2022bailando,zhang2023t2m,tevet2022motionclip,temos-eccv-2022}.

Skeleton sequences from each datasets are pre-processed in the same way. We filtered out skeletons that have too small or too large joint variance, which corresponds to static or erratic movement, respectively. We found that this procedure ensures that only properly moving skeletons are kept in the dataset. Furthermore, skeleton sequences are normalized and aligned at the pelvis, using the following formulae, considering that each of the $J = 18$ joints have $(x, y)$ coordinates:

\begin{equation*}
    \hat{x}_{joint} = \frac{x_{joint} - x_{pelvis}}{|x_{R.shoulder} - x_{L.shoulder}|}
    \label{eq:eq1}
\end{equation*}
\begin{equation*}
    \hat{y}_{joint} = \frac{y_{joint} - y_{pelvis}}{|y_{neck} - y_{pelvis}|}
    \label{eq:eq2}
\end{equation*}

\begin{table}[hbt!]
    \centering
    \resizebox{0.85\linewidth}{!}{
        \begin{tabular}{c|ccc}
            \textbf{Dataset} & \textbf{Split} & \textbf{\# Sequences} & \textbf{Duration (hr.)}\\
            \midrule\midrule
            \multirow{2}{*}{DenseGait \cite{cosma22gaitformer}} & Train & 217,954 & 614.75\\
            & Validation & 10,733 & 36.53 \\
            \midrule
    
            \multirow{3}{*}{GREW \cite{grew}} & Train & 102,888 & 175.92 \\
            & Test & 24,000 & 65.37\\
            & Distractor & 226,588 & 154.82\\
            \midrule
    
            \multirow{2}{*}{OU-ISIR \cite{ouisir}} & Train & 133,872 & 57.82 \\
            & Test & 134,199 & 57.92 \\
            \midrule
    
            \multirow{2}{*}{Gait3D \cite{zheng2022gait3d}} & Train & 18,940 & 42.70 \\
            & Test & 6,369 & 14.23 \\
            \midrule
    
            \textbf{Total} & & \textbf{875,543} &  \textbf{1220.06} \\
            
        \end{tabular}
    }
    \caption{Datasets that make up our pretraining dataset. We combined all the major in-the-wild and controlled datasets (including all splits) into a single, large-scale and diverse dataset. The dataset contains gait samples from a diverse set of walking registers, environments and camera angles. }
    \label{tab:walkpile}
\end{table}

In this manner, every skeleton sequence is aligned spatially and the differences in height and width of individuals are essentially eliminated. Consequently, only movement is encoded irrespective of the screen coordinates, distance to camera or appearance cues. We employ minimal augmentations to the skeleton sequences, adopting only random temporal cropping and walking pace modifications \cite{9721551,wang2020self,cosma22gaitformer}. We crop each skeleton sequence to be $T = 64$ frames long.

\begin{figure*}
    \centering
    \includegraphics[width=0.75\textwidth]{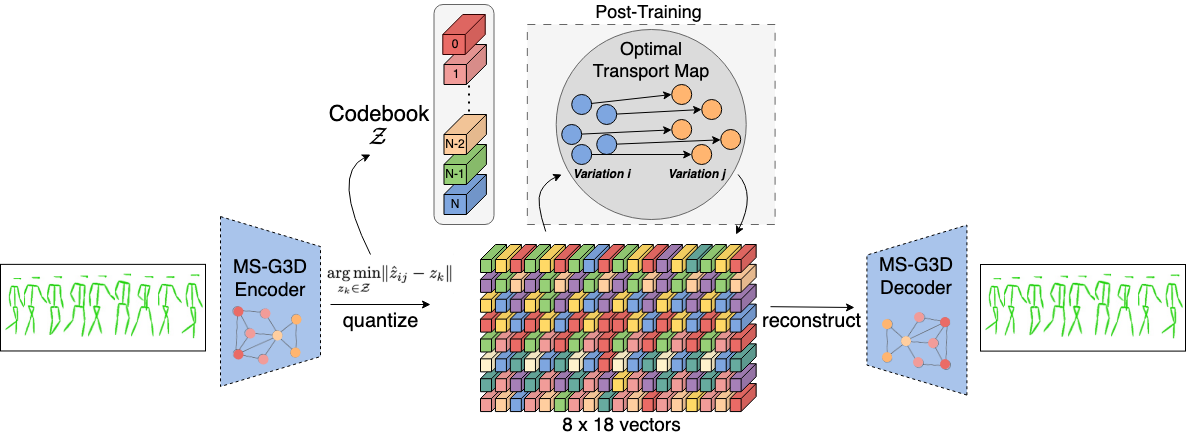}
    \caption{Overall architecture of GaitMorph. We train a MS-G3D encoder-decoder to quantize gait representations into a learned fixed-size codebook. After training, we can manipulate the discrete latent space and morph a walking variation into another using a transport map learned on the training set of a controlled walking dataset.}
    \label{fig:architecture}
\end{figure*}

\subsection{Learning the Walking Codebook}

In order to learn an informative and context-rich walking codebook, we leverage the expressive power of a Vector Quantized Variational AutoEncoder model (VQ-VAE) \cite{van2017neural}. The VQ-VAE model has been shown to be effective for a range of tasks, including image compression and generation \cite{esser2020taming,razavi2019generating}, and speech recognition \cite{van2017neural}. It is particularly useful in situations where the input data has a high degree of variability, and where traditional continuous latent space models may struggle to capture the underlying structure of the data. Furthermore, a discrete latent space enables a high degree of data compression, and allows the input data to be further processed as a sequence of discrete tokens. 

To properly encode skeleton sequences, we construct a skeleton autoencoder based on the MS-G3D \cite{liu2020disentangling} model. Figure \ref{fig:architecture} showcases the overall architecture of our method. MS-G3D is a powerful graph convolutional model that has state-of-the-art results in skeleton action recognition, surpassing other graph-based methods \cite{yan2018spatial,shi2019two} by a large margin. Graph convolutional models are well established in the field of skeleton sequence processing \cite{gupta2021quo} and were developed to properly handle spatial and temporal variation of the skeleton graph.

\subsubsection{MS-G3D Encoder-Decoder}

The encoder and decoder models for the skeleton autoencoder are both based on the MS-G3D architecture \cite{liu2020disentangling}. For simplicity, we did not perform any graph subsampling \cite{yan2019convolutional}, and only used temporal pooling to compress the skeleton sequence. We follow the official model implementation \cite{liu2020disentangling}, and adapt it for gait processing. Specifically, we changed all activations to GeLU \cite{hendrycks2016gaussian}, we removed the initial data batch-normalization since the skeletons were already normalized. Initial experiments showed that the default model was not large enough to reconstruct sequences other than the mean skeleton. Consequently, we doubled each convolution - batch normalization - activation block to increase model capacity. A MS-G3D model is composed of multiple Spatial-Temporal Graph Convolution (ST-GC) blocks. Each block consists of a Multi-scale Graph Convolution block (MS-GCN) and two Multi-Scale Temporal Convolutional blocks \cite{liu2020disentangling}. We used the default 6 G3D scales and 13 GCN scales for both the encoder and decoder models.

For the MS-G3D encoder $E(\cdot)$, we used 20 ST-GC encoder blocks. We used a feature map size of 64 for the first 5 blocks, 128 for the next 5 and 256 for the final 10. Temporal pooling is performed every 5 blocks. Therefore, for an initial skeleton sequence $x \in \mathbb{R}^{T \times J \times 2}$ consisting of $T = 64$ skeletons (i.e. frames) with $J = 18$ joints, the sequence is temporally downsampled to $\hat{z} \in \mathbb{R}^{\frac{T}{4} \times J \times n_{\hat{z}}}$, where, in our case, $n_{\hat{z}} = 256$ the encoder embedding size. 

For the MS-G3D decoder $G(\cdot)$, we opted for a slightly smaller model, since we experimentally observed that the encoder size is more negatively correlated to the final reconstruction error than the decoder size. Moreover, having a smaller decoder is more computationally efficient, and enables faster reconstruction of latent codes. The overall constituent decoder blocks are identical to the encoder blocks, but we replaced the strided convolution with a strided deconvolutional block for the temporal upsampling. We used 16 ST-GC blocks, with feature maps of size 32 for the first 4 blocks, 16 for the next 4 and 8 for the final 8. Temporal upsampling was performed every 4 blocks. 

\subsubsection{Skeleton Vector Quantization}

Instead of utilizing a continuous latent space to encode the skeleton sequences, we quantize each latent embedding into a fixed-length learnable codebook $\mathcal{Z} = \{z_k\}^K_{k=1} \subset \mathbb{R}^{n_z}$. Any skeleton sequence $x \in \mathbb{R}^{T \times J \times 2}$ is encoded using the MS-G3D encoder described above into a temporally-compressed representation $\hat{z} \in \mathbb{R}^{\frac{T}{4} \times J \times n_{\hat{z}}}$, which is then quantized into $z_{\textbf{q}} \in \mathbb{R}^{\frac{T}{4} \times J \times n_{z}}$, where $n_{\textbf{z}}$ is the codebook dimensionality, not necessarily equal to the encoder embedding size. Each $\hat{z}$ is encoded using a nearest neighbor search in the codebook (see Eq. \ref{eq:quantization-nn}).

\begin{equation}
    z_{\mathbf{q}} = \mathbf{q}(\hat{z}) \coloneqq \argmin_{z_k \in \mathcal{Z}} \lVert \hat{z}_{ij} - z_k \rVert \in \mathbb{R}^{T \times J \times n_z}
    \label{eq:quantization-nn}
\end{equation}

After quantization, skeletons are reconstructed using the MS-G3D decoder: $\hat{x} = G(z_{\textbf{q}}) = G(\textbf{q}(E(x)))$. The model is trained end-to-end using a stop-gradient operation (see Eq. \ref{eq:vq-loss}) since the dictionary look-up is not differentiable. For more details regarding training, readers are referred to the work of Van Den Oord et al. \cite{van2017neural}.

\begin{equation}
\begin{split}
    \mathcal{L}_{VQ} (E, G, \mathcal{Z}) = \lVert x - \hat{x} \rVert &+ \lVert \text{sg}[E(x)] - z_{\mathbf{q}} \rVert_2^2 \\
    & + \lVert \text{sg}[z_{\mathbf{q}}] - E(x) \rVert_2^2
\end{split}
\label{eq:vq-loss}
\end{equation}

In practice, instead of the $l_2$ loss for the reconstruction error, we employed a Smooth $l_1$ with $\beta = 0.25$ \cite{girshick2015fast} to further penalize small reconstruction errors. VQ-VAE models are notoriously hard to train \cite{lancucki2020robust}, primarily due to the dictionary collapse problem, in which most of the codebook entries are not utilized in reconstruction, yielding poor performance. To deal with this problem, we employed a standard array of "bag-of-tricks" to increase codebook usage. We used K-means initialization of the codebook \cite{zeghidour2021soundstream}, we used a lower codebook dimensionality of $n_z = 16$ by linearly projecting down the encoder embedding, we used cosine similarity for codebook search \cite{yu2021vector}, expiring stale codes \cite{zeghidour2021soundstream} and orthogonal regularization \cite{shin2021translation} of the codebook vectors to encourage linear independence. The codebook is learned using an exponential moving average approach with a decay rate of $\gamma = 0.9$. Autoencoder warm-up \cite{fu-etal-2019-cyclical} was not necessary. We experimented with using a separate codebook for each limb, similar to Xie et al. \cite{xie2022vector}, but did not observe a substantial improvement.

Training was performed on a single NVIDIA RTX 3060, using mixed-precision training, with a batch size of 48. The network was updated for 50k steps, using AdamW \cite{kingma2014adam} optimizer using a cyclical learning rate schedule \cite{cyclical-lr} which varies the learning rate between 0.0025 and 0.0075. The model has 4.8M non-embedding parameters. The training duration for the VQVAE is approximately 15 hours.

\subsection{Learning Optimal Transport Mappings}

In order to exploit the expressive power of the learned gait tokens, we posit that only specific tokens from a tokenized gait sequence are responsible for encoding the gait viewpoint and variation. Therefore, for a set of walks from a particular variation $\mathcal{T}$, we can learn a set of transport maps $\Gamma = \{\gamma_j^{\textbf{*}} | j \in 1 \dots (\frac{T}{4} \times J)\}$, for each encoded position $j$, that transform the target quantized gait representation into a quantized representation of a baseline walk $\mathcal{B}$. The transformed walk $\mathcal{T}$ is then decoded by the generator: $\mathcal{T}^* = G(\Gamma(\textbf{q}(E(\mathcal{T}))))$. The walks $\mathcal{B}$ and $\mathcal{T}^*$ should be from the same walking variation.  We propose to learn the transport maps $\Gamma$ by utilizing optimal transport theory \cite{villani2009optimal}. We learn a transport map $\gamma^*_j$ by minimizing the Earth Mover's Distance (EMD) between the histograms of two quantized gaits. EMD assumes there is a cost for moving one quantity to another, which is encoded into a cost matrix $C$. In general, EMD is defined as:

\begin{equation}
\begin{split}
    \gamma^* = \argmin_{\gamma \in \mathbb{R}_{+}^{m \times n}} \sum_{i,j} \gamma_{i,j} C_{i,j} \\
    \text{s.t.} \gamma 1 = a; \gamma^T 1 = b; \gamma \geq 0
\end{split}
\label{eq:ot}
\end{equation}

In our case, $a$ and $b$ are histograms of the token occurrences in each gait sequence, and the cost matrix $C$ is given by the pairwise distances between the token embeddings. To account for multiple occurrence of the same token in a quantized gait sequence, we scale the corresponding vector embedding by the number of occurrences. We describe our method in Algorithm \ref{alg:gait-morphing}. The algorithm is an instance of an assignment problem for each token position, and is similar to finding the minimum flow between the two token distributions. In practice, we use the algorithm proposed by Bonneel et al. \cite{bonneel2011displacement} implemented in the PyOT \cite{flamary2021pot} library.

\begin{algorithm}
    \caption{Finding the optimal transport maps between walking variations. } 
    \label{alg:gait-morphing}
    \begin{algorithmic}
     \scriptsize
        \Require \\
        $E$ - Trained MS-G3D gait encoder \\
        $\mathcal{B} \in \mathbb{R}^{B^{(b)} \times T \times J \times 2}$ - baseline variation walks \\
        $\mathcal{T} \in \mathbb{R}^{B^{(t)} \times T \times J \times 2}$ - target walks \\
        $\mathcal{Z}$ - learned codebook vectors \\
        $s$ - token sequence length\\
        
        \State $k^{(b)} \gets \arg (\textbf{q}(E(\mathcal{B})))$ \Comment{\textit{Baseline token indices.}}
        \State $k^{(t)} \gets \arg (\textbf{q}(E(\mathcal{T})))$ \Comment{\textit{Target token indices.}}

        \State $\Gamma \gets \emptyset$
        \For{$j \gets 1 \dots s$} 
            \LeftComment \textit{Count occurrences of each baseline and target tokens.}

            \State $c^{(b)} \gets \{\sum_l^{B^{(b)}} \mathbb{1}[k^{(b)}_{l, j} = r] | r \in 1 \dots |\mathcal{Z}|\}$
            \State $c^{(t)} \gets \{\sum_l^{B^{(t)}} \mathbb{1}[k^{(t)}_{l, j} = r] | r \in 1 \dots |\mathcal{Z}|\}$

            \LeftComment \textit{Increase codebook embedding magnitude.}
            \State $C^{(b)} \gets \mathcal{Z} \odot c^{(b)}$
            \State $C^{(t)} \gets \mathcal{Z} \odot c^{(t)}$

            \LeftComment \textit{Compute cost matrix as pairwise distances between scaled token embeddings.}
            \State $C \gets C^{(b)} \cdot (C^{(t)})^\top$
            
            \LeftComment \textit{Find optimal transport map for position \textit{j}}
            \State $\gamma^{\textbf{*}}$ $\gets$ $\argmin_{\gamma} \sum \gamma C$ \Comment{Eq. \ref{eq:ot}}
            \State $\Gamma_j \gets \gamma^{\textbf{*}}$ 
        \EndFor
        
        \State \textbf{return} $\Gamma$
    \end{algorithmic}
\end{algorithm}

In practical scenarios where the gait variation is not known beforehand, domain-expert models such as pedestrian attribute identification models \cite{wang2022pedestrian} can be used to estimate particular walking attributes, similar to the approach of Cosma and Radoi \cite{cosma22gaitformer}, to inform the morphing target. This method can also be used as data augmentation to generate multiple views for the same walking sequence, for use in contrastive self-supervised training \cite{jaiswal2020survey,cosma22gaitformer}.

\section{Experiments and Results}
\begin{figure*}[hbt!]
    \centering
    \includegraphics[width=0.85\linewidth]{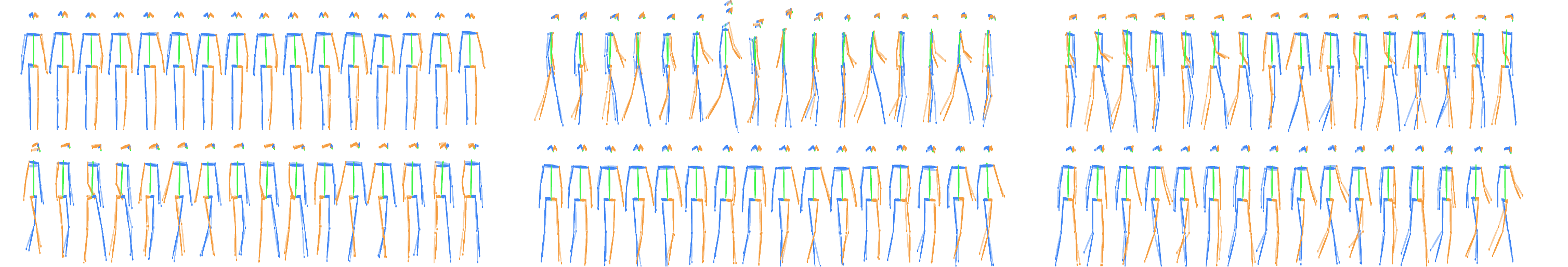}
    \caption{Overlapped original and reconstructed gait sequences from CASIA-B validation set using a VQ-VAE with $|\mathcal{Z}| = 8$. We differentiate {\color{orange} left} and {\color{blue} right} laterals with appropriate colors. Original skeletons are transparent, while reconstructed skeletons are opaque. The model can properly reconstruct sequences, and acts as a low-pass filter on the skeleton sequence, dampening exaggerated movements caused by inaccurate pose extraction (middle-top). Best viewed in color.}
    \label{fig:recons}
\end{figure*}

For our experiments, we used CASIA-B \cite{casia} and Front-View Gait (FVG) \cite{fvg} to evaluate the performance of our proposed method. CASIA-B is a popular gait recognition dataset, widely used to test the robustness of gait analysis model across multiple viewpoints and walking variations. It contains gait sequences from 124 subjects, captured in 11 different viewpoints, under three walking variations: normal walking (NM), clothing walking (CL) and walking with a bag (BG). For a walker, 6 sessions are captured under the normal walking variation, 2 sessions with clothing change and 2 sessions under the bag carrying variation. Each walk has its variation / viewpoint known. We use the standard \cite{casia} training / testing split, consisting of the first 62 training subjects, with all available walking sessions. FVG is another popular dataset for gait recognition that features walks from only the front facing viewpoint, which is considered the most challenging due to the reduced perceived movement variation. It is comprised of 226 subjects captured in 6 walking variations: normal walk (NM), walk speed (WS), change in clothing (CL), carrying bag (CB), cluttered background (CBG) and ALL. In our experiments, we omit the "ALL" variation to properly isolate confounding factors. We used the first 136 subjects as the training split, and the rest for testing. It is important to note that the VQ-VAE model is trained on the dataset described in Section \ref{sec:dataset}, and remains frozen throughout the rest of the experiments. Moreover, the transport maps are learned only on the training split of each dataset (CASIA-B / FVG) and are utilized as-is on the testing split.

\subsection{The effect of dictionary size on the reconstructed gait sequences}

\begin{figure}[hbt!]
    \centering
    \includegraphics[width=0.65\linewidth]{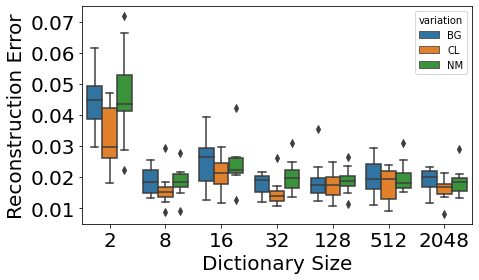}
    \caption{Boxplots for the across viewpoint distribution of reconstruction errors for different dictionary sizes.}
    \label{fig:casia-recons}
\end{figure}

\begin{figure}[hbt!]
    \centering
    \includegraphics[width=0.65\linewidth]{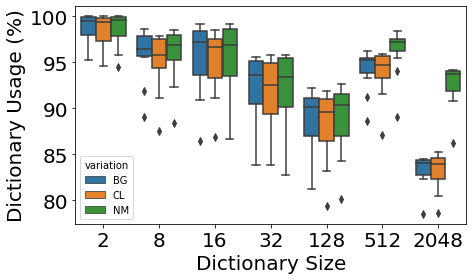}
    \caption{Dictionary usage for each of the trained VQ-VAE models. Increasing the dictionary size is correlated with lower dictionary usage and codebook under-utilization.}
    \label{fig:casia-usage}
\end{figure}

We trained several VQ-VAE models with increasingly larger dictionary sizes ($|\mathcal{Z}| \in \{2, 8, 16, 32, 128, 512, 2048\}$) to gauge the effect on the reconstructed sequences. In Figure \ref{fig:casia-recons}, we showcase the reconstruction error for each model on CASIA-B evaluation set, for each walking variation. For $|\mathcal{Z}| = 2$, the results are significantly worse than other dictionary sizes, due to the extreme compression. The model with $|\mathcal{Z}| = 8$ achieves the best overall performance. We showcase qualitative reconstruction samples in Figure \ref{fig:recons} - the model can reliably reconstruct skeleton sequences and acts as a low-pass filter which dampens exaggerated movements caused by inaccurate pose extractions.

Our model achieves a high degree of compression for skeleton sequences - a skeleton sequence represented as a float32 sequence of 2304 numbers is equivalent to storing 73728 bits of information, but using a VQ-VAE approach, the storage space is reduced to only 144 bits for $|\mathcal{Z}| = 2$ and 432 bits for $|\mathcal{Z}| = 8$. This compression level potentially allows on-device storage of massive amounts of skeleton sequences. 

Figure \ref{fig:casia-usage} showcases the dictionary usage for each dictionary size. Increasing the dictionary size slightly decreases dictionary usage, which implies that some tokens are underutilized by the model. This effect is more pronounced for $|\mathcal{Z}| = 2048$, especially for non-normal walking variations. This is most likely due to the fact that the pretraining dataset for the VQ-VAE mostly contains in-the-wild walks, which make the BG and CL variations easier to reconstruct.  The reconstructed gait sequences are not detrimental to downstream gait recognition models. To gauge the faithfulness of the reconstructed skeletons to the real walking skeletons, in Table \ref{tab:recog-recons} we showcase gait recognition results for CASIA-B using reconstructed skeletons as training data. The performance loss by using reconstructed skeletons is marginal, and even beneficial in some cases. This result can be attributed to the fact that the VQ-VAE acts as a low-pass filter and can slightly improve data quality across training. Furthermore, performance on gait recognition is not correlated with the reconstruction error of the VQ-VAE: the model with $|\mathcal{Z}| = 2$ achieves comparable results with the baseline method using real skeletons.

\begin{table}[hbt!]
    \centering
    \resizebox{0.85\linewidth}{!}{
        \begin{tabular}{llll}
         &           \textbf{NM} &           \textbf{BG} &           \textbf{CL} \\
        \midrule
        Baseline                &  0.79 ± 0.06 &  0.46 ± 0.08 &  0.24 ± 0.06 \\
        \midrule
        $|\mathcal{Z}| = 2$  &   0.76 ± 0.09 &  0.46 ± 0.09 &  0.24 ± 0.07 \\
        $|\mathcal{Z}| = 8$     &  0.75 ± 0.13 &  0.44 ± 0.08 &  0.21 ± 0.08 \\
        $|\mathcal{Z}| = 16$    &  0.75 ± 0.12 &  0.44 ± 0.08 &  0.22 ± 0.08 \\
        $|\mathcal{Z}| = 32$    &  0.76 ± 0.12 &   0.47 ± 0.1 &  0.23 ± 0.09 \\
        $|\mathcal{Z}| = 128$   &   0.78 ± 0.1 &  0.46 ± 0.09 &  0.22 ± 0.07 \\
        $|\mathcal{Z}| = 512$   &  0.76 ± 0.12 &  0.46 ± 0.09 &  0.22 ± 0.09 \\
        $|\mathcal{Z}| = 2048$ &  0.78 ± 0.11 &  0.47 ± 0.11 &   0.24 ± 0.1 \\
        \end{tabular}
    }
    \caption{Accuracy on CASIA-B for GaitFormer \cite{cosma22gaitformer} trained with reconstructed skeletons. We report the mean and standard deviation of recognition accuracy across 4 distinct runs and all viewpoints.}
    \label{tab:recog-recons}
\end{table}

\subsection{Evaluation of morphed gait sequences}

\begin{figure*}[hbt!]
    \centering
    \includegraphics[width=0.80\linewidth]{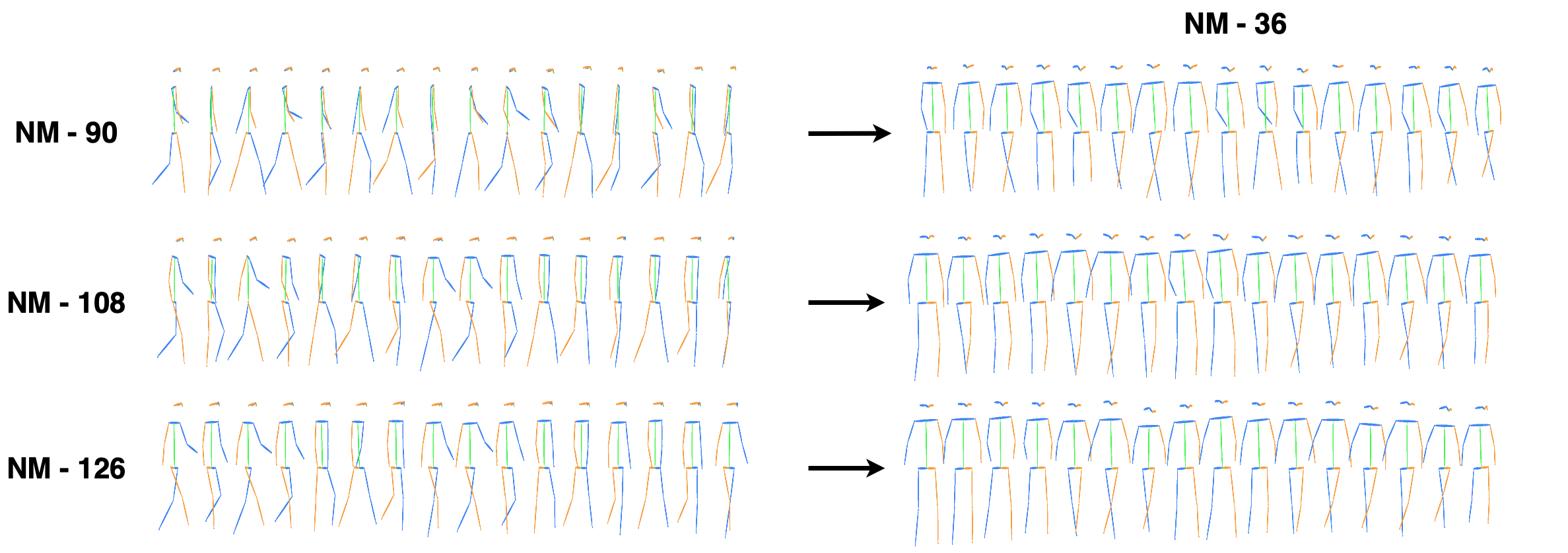}
    \caption{Examples of modified skeleton sequences using optimal transport maps. We differentiate {\color{orange} left} and {\color{blue} right} laterals with appropriate colors. The model is able to successfully change the walking viewpoint to a normal walk under viewpoint 36$^\circ$ (NM-36). For this example, we chose a VQ-VAE with $|\mathcal{Z}| = 512$. Best viewed in color. }
    \label{fig:casia-morphs}
\end{figure*}

\begin{figure*}[hbt!]
    \centering
    \includegraphics[width=0.60\linewidth]{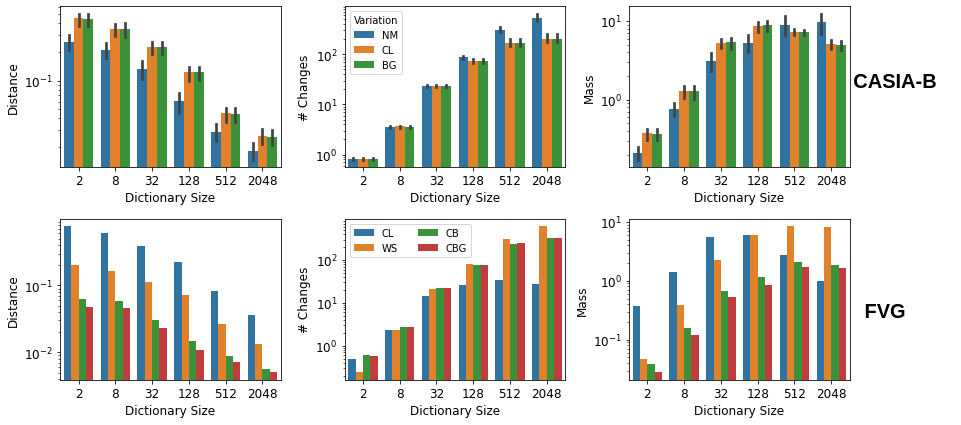}
    \caption{The average moved distance, number of changes and total mass moved between variations for CASIA-B and FVG. In the case of CASIA-B, the error bars represent standard deviation across viewpoints. }
    \label{fig:mass}
\end{figure*}

We first present a qualitative evaluation for gait morphing. Figure \ref{fig:casia-morphs} showcases selected gait sequences from three different viewpoints morphed to a common NM-36 variation. The model is able to morph sequences into the baseline sequence, properly handling limb switching (left and right limbs are properly swapped when the viewpoint is from behind the walker). For similar baseline / target pairs, the transport maps exhibit fewer changes. Transport maps from VQ-VAE models with a larger dictionary size exhibit more token changes, but the earth movers distance between variations is comparatively smaller. This implies that many smaller changes are performed with same effect. In Figure \ref{fig:mass} we showcase the average moved distance, the number of changes and the total mass moved across the walking variations for both CASIA-B and FVG. We define mass as the number of changes multiplied by the average change cost. The number of changes is larger when the dictionary size is larger, but the total mass moved remains constant after the $|\mathcal{Z}| = 128$. This implies that the tokens from the model with $|\mathcal{Z}| = 2048$ are more disentangled, since less mass is moved to achieve the same outcome. 

In terms of numeric evaluation, our goal is to compare the morphed walks to the real walks of a particular variation. In our experiments, our comparisons are made with regard to NM-36 variation for CASIA-B and NM for FVG. The most straightforward comparison is to use mean squared error between skeletons, but we have no guarantees of sequence alignment between variations in either dataset. As such, we propose a metric between walking distributions, similar to the FID \cite{heusel2017gans} distance. The Frechet Inception Distance (FID) was introduced by Heusel et al. \cite{heusel2017gans} to measure the generation quality of GANs compared to real images. The FID score is based on the Frechet Distance \cite{dowson1982frechet}, and measures the distance $d(\cdot)$ between two gaussian distributions $\phi = (\textbf{m}, \textbf{C})$ and $\phi_w = (\textbf{m}_w, \textbf{C}_w)$, corresponding to a distribution of real and synthetic samples, respectively: $d^2(\phi, \phi_w) = ||\textbf{m} - \textbf{m}_w||^2 + Tr(\textbf{C} + \textbf{C}_w - 2(\textbf{CC}_w)^{\frac{1}{2}})$. The means and standard deviations of the Gaussians are, for images, the means and standard deviations of a set of embedding vectors of an Inception network \cite{inception} pretrained on ImageNet. The metric captures levels of perceived disturbance between real and synthetic samples \cite{heusel2017gans}. 

For gait synthetisation, we propose a specialized variant of the FID score, which we name "\textit{Frechet Gait Distance (FGD)}", in which walks are processed by a pretrained GaitFormer network on DenseGait \cite{cosma22gaitformer}. FGD stands as a automatic measure of walking "naturalness", by measuring the similarity to a given real gait distribution. Variants have been proposed for measuring motion naturalness and are geared towards general action synthesis \cite{gopinath2020fairmotion,siyao2022bailando,maiorca2022evaluating}, but a specialized variant for gait has not yet been adopted.

\begin{table}[]
    \centering
    \resizebox{\linewidth}{!}{
        \begin{tabular}{l|lllllll}
             &  &      \textbf{0$^\circ$} &     \textbf{72$^\circ$} &     \textbf{90$^\circ$} &    \textbf{126$^\circ$} &    \textbf{162$^\circ$} &   \textbf{ 180$^\circ$} \\
            \midrule
\multirow{8}{*}{\textbf{NM}} &  \textit{Baseline (vs real NM-36)} &  \textit{\textbf{0.045532}} &  \textit{\underline{0.070282}} &  \textit{\underline{0.111757}} &  \textit{0.138415} &   \textit{0.19525} &  \textit{0.265378} \\
             & \textit{Heuristic Aug. (vs real NM-36)} &  0.047659 &  0.076971 &  0.115972 &  0.138536 &  0.195943 &   0.27324 \\
             &  $|\mathcal{Z}| = $ 2 &  0.754251 &  0.320832 &  0.315892 &  0.195995 &  0.339743 &   0.74698 \\
             & $|\mathcal{Z}| = $ 8 &  0.379271 &  0.200036 &  0.104054 &  0.207928 &  0.673472 &  0.549586 \\
             & $|\mathcal{Z}| = $ 16 &  0.136429 &   0.11086 &   0.22199 &  0.434574 &  0.530231 &  0.128465 \\
             & $|\mathcal{Z}| = $ 32 &  0.091645 &  0.293709 &  0.400562 &  0.557834 &   0.63797 &   0.20792 \\
            & $|\mathcal{Z}| = $ 128 &   0.08184 &  0.465142 &  0.597904 &  0.766841 &  0.697738 &  0.268043 \\
            & $|\mathcal{Z}| = $ 512 &  0.074983 &   0.11027 &  0.117966 &  \underline{0.110944} &  \underline{0.140733} &  \textbf{0.107217} \\
           & $|\mathcal{Z}| = $ 2048 &  \underline{0.046048} &  \textbf{0.060231} &  \textbf{0.082002} &  \textbf{0.102774} & \textbf{0.104749} &  \underline{0.135883} \\
            \midrule
  \multirow{8}{*}{\textbf{BG}} & \textit{Baseline (vs real NM-36)} &  \textit{\textbf{ 0.05295}} &  \textit{\underline{0.074746}} &  \textit{0.114694} & \textit{ 0.150358} &  \textit{0.211948} &  \textit{0.274384} \\
               & \textit{Heuristic Aug. (vs real NM-36)} &  0.055826 &  0.083356 &  0.119362 &  0.152413 &  0.209289 &  0.283982 \\
              & $|\mathcal{Z}| = $ 2 &  0.716497 &  0.304378 &  0.243575 &  0.193439 &  0.599968 &  0.743501 \\
              & $|\mathcal{Z}| = $ 8 &  0.320088 &  0.184071 &  0.177406 &  0.190533 &  0.653437 &  0.639077 \\
             & $|\mathcal{Z}| = $ 16 &  0.169745 &  0.129582 &  0.232889 &  0.455946 &  0.536296 &  0.185364 \\
             & $|\mathcal{Z}| = $ 32 &  0.088956 &  0.341674 &  0.407447 &  0.589189 &  0.635381 &  0.203955 \\
            & $|\mathcal{Z}| = $ 128 &   0.07735 &  0.343908 &  0.514205 &  0.592576 &  0.527332 &  0.307418 \\
            & $|\mathcal{Z}| = $ 512 &  0.080196 &  \textbf{0.062556} &  \textbf{0.070378} &  \textbf{0.094266} &  \textbf{0.115324} &  \textbf{0.136357} \\
           & $|\mathcal{Z}| = $ 2048 &  \underline{0.056126} &  0.081991 &  \underline{0.106456} &  \underline{0.131166} &  \underline{0.137103} &  \underline{0.161214} \\
            \midrule
  \multirow{8}{*}{\textbf{CL}} & \textit{Baseline (vs real NM-36)} & \textit{ 0.110895} &  \textit{0.140185} &  \textit{0.189128} &  \textit{0.230226} &  \textit{0.320092} &  \textit{0.411968} \\
            & \textit{Heuristic Aug. (vs real NM-36)} &  0.120972 &  0.147726 &  0.197784 &  0.235666 &  0.318236 &  0.420584 \\
              & $|\mathcal{Z}| = $ 2 &  0.726999 &  0.312846 &   0.30105 &  0.376383 &  0.338582 &  0.670051 \\
              & $|\mathcal{Z}| = $ 8 &  0.261182 &  0.191699 &  0.129651 &  0.219145 &  0.656104 &  0.521065 \\
             & $|\mathcal{Z}| = $ 16 &  0.142326 &  0.194611 &  0.273543 &   0.50944 &  0.566114 &  0.177739 \\
             & $|\mathcal{Z}| = $ 32 &   0.07656 &  0.316372 &  0.418504 &  0.572725 &  0.589525 &  0.217498 \\
            & $|\mathcal{Z}| = $ 128 &  \textbf{0.064639} &  0.380276 &  0.514381 &  0.531558 &    0.4364 &  0.284238 \\
            & $|\mathcal{Z}| = $ 512 &  0.084125 &  \textbf{0.057824} &  \textbf{0.063853} &  \textbf{0.095734} &  \textbf{0.128801} &  \textbf{0.147653} \\
           & $|\mathcal{Z}| = $ 2048 &  \underline{0.075194} &  \underline{0.096743} &  \underline{0.128168} &  \underline{0.148594} &  \underline{0.159419} &  \underline{0.192654} \\
        \end{tabular}
    }
    \caption{FGD values between the morphed gait to the NM-36 variation and the real NM-36 for CASIA-B validation set. Baseline values corresponds to the FGD between the real unmodified gait and NM-36. In most variations, the morphed walk is much closer to the real NM-36 than the unmodified walk, especially for extreme viewpoints. We denote with \textbf{bold} the smallest distance and with \underline{underline} the second smallest distance.}
    \label{tab:fgd-casia}
\end{table}

\begin{table}[]
    \centering
    \resizebox{\linewidth}{!}{   
        \begin{tabular}{l|llll}
         & \textbf{WS} &        \textbf{CB} &        \textbf{CL} &       \textbf{CBG} \\
        \midrule
                      \textit{Baseline (vs real NM)} &  \textit{\textbf{0.001754}} & \textit{\textbf{ 0.039509}} &  \textit{\textbf{0.014785}} &  \textit{\textbf{0.001582}} \\
                      \textit{Heuristic Aug. (vs real NM-36)}  &  0.002493 &  0.042682 &   0.01474 &  0.001812 \\
                      $|\mathcal{Z}| = $ 2 &  0.051189 &  0.081181 &  0.189054 &  0.077997 \\
                      $|\mathcal{Z}| = $ 8 &  0.055022 &  0.100748 &  0.107133 &  0.082222 \\
                     $|\mathcal{Z}| = $ 16 &  0.032046 &   0.04954 &  0.067667 &  0.053239 \\
                     $|\mathcal{Z}| = $ 32 &  0.031136 &  0.058868 &  0.059492 &   0.04834 \\
                    $|\mathcal{Z}| = $ 128 &  0.032589 &  0.059443 &  0.056675 &  0.034228 \\
                   $|\mathcal{Z}| = $ 512 &   0.03081 &  0.050809 &  0.035253 &  0.028172 \\
                   $|\mathcal{Z}| = $ 2048 &  \underline{0.022087} &  \underline{0.043005} &  \underline{0.019479} &  \underline{0.025608} \\
        \end{tabular}
}
    \caption{FGD values between the morphed gait to the NM variation and the real NM for FVG validation set. Baseline values corresponds to the FGD between the real unmodified gait and NM. The morphed walk is similar to the real NM variations, but the effect is not pronounced due to the same underlying viewpoint for all variations. We denote with \textbf{bold} the smallest distance and with \underline{underline} the second smallest distance. 
    }
    \label{tab:fgd-fvg}
\end{table}

In Tables \ref{tab:fgd-casia} and \ref{tab:fgd-fvg}, we present our results for gait morphing for CASIA-B and FVG, respectively. We utilized the proposed FGD metric to compare the distance between the distribution of the morphed walks to the real baseline walking variation (NM-36 for CASIA-B and NM for FVG). For CASIA-B we focus our evaluation in terms of viewpoint, since it is the principal confounding factor, especially for 2D poses. Results show that the morphed walks are properly generated and are closer to the real NM-36 walking variation compared to the unmodified walk and for more extreme viewpoints, the effect is larger. Results are more correlated with the dictionary usage for each dictionary size, rather than reconstruction error (which is low for every dictionary size). Additionally, we compared morphed gaits with standard array of heuristic skeleton augmentations present in other works\cite{cosma22gaitformer,gaitgraph}: random pace with a time multiplier sampled from \{0.5, 0.75, 1, 1.25, 1.5, 1.75, 2.0\}, joint and point noise with standard deviation of 0.001, random mirroring and reversing the walk. While heuristic augmentations provide some variation in the vicinity of the original walk, the FGD across views are similar to the non-augmented walks. These results show that the morphed walks with our method represent a good way to augment existing walks to synthesize novel views. Since all the walks in FVG are from the same viewpoint, the differences between walking variations are not as evident. Consequently, the distance between distributions is comparatively smaller than in CASIA-B. 

\begin{figure}[hbt!]
    \centering
    \includegraphics[width=0.45\linewidth]{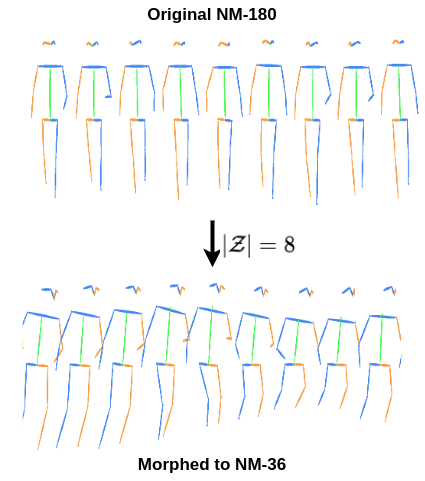}
    \caption{Failure case for $|\mathcal{Z}|=8$ when morphing a normal walk from CASIA-B from viewpoint 180$^\circ$ to viewpoint 36$^\circ$. The latent space is not sufficiently disentangled to learn a general transport map without severely distorting the resulting gait sequence.}
    \label{fig:failure}
\end{figure}

It is clear from results in Tables \ref{tab:fgd-casia} and \ref{tab:fgd-fvg} that models operating with a low dictionary size are not appropriate to be used for morphing. This is most likely due to the latent embeddings being severely entangled. Figure \ref{fig:failure} showcases a selected failure case for morphing a NM-180 walk from CASIA-B into NM-36 using a VQ-VAE with $|\mathcal{Z}| = 8$. The generated walk has severe artifacts and cannot be considered appropriate for downstream model training. Inherently, there is a trade-off between dictionary size and the manipulability of the latent codes: larger dictionary sizes have more disentangled representations which allow for more informed changes at the expense of lower data compression.

\section{Conclusions}
In this work, we presented GaitMorph, a novel method for modifying gait sequences into new walking variations. Our proposed approach entails firstly training a discrete latent model (in our case, a VQ-VAE) that compresses the walking sequences into a sequence of interpretable tokens, and learning an optimal latent transport map across variations. Our extensive experiments show that the trained VQ-VAE model preserves the walker's identity, achieving a marginal loss in performance when utilizing reconstructed sequences in gait recognition scenarios. Furthermore, we showed that the distribution of morphed sequences is similar to the real walk distribution. 
Our approach has the potential to be applied to self-supervised learning scenarios for gait recognition \cite{cosma22gaitformer}, which are heavily reliant \cite{chen2020simple,tian2020makes} on multiple strong augmentations / views for the same input.

{\small
\bibliographystyle{ieee}
\bibliography{egbib}
}

\end{document}